\documentclass{gvdb-sig-alternate}

\usepackage{microtype}
\usepackage{url}

\usepackage{latexsym}
\usepackage{booktabs}
\usepackage{acronym}
\usepackage{standalone}
\usepackage{subfig}
\usepackage{forest}
\usepackage{xcolor}
\usepackage{pgfplots}
\usepackage{pgfplotstable}
\usepackage{float}
\usepackage{listings}
\usepackage{multirow}
\usepackage{mathabx}
\usepackage{tikz}
\usetikzlibrary{arrows.meta,calc,pgfplots.groupplots,positioning}
\usepackage{adjustbox}
\usepackage{array}
\usepackage{xfrac}
\newcolumntype{R}{%
    >{\adjustbox{angle=90,lap=\width-(1em)}\bgroup}%
    l%
    <{\egroup}%
}

\usepgfplotslibrary{colorbrewer}

\tikzset{
  sibling distance=2em, 
  level distance=2em, 
  tg/.style={
    draw,
    circle,
    fill=white,
    text width=0.7em,
    inner sep=0em
  }
}

\definecolor{p}{RGB}{255,80,80}
\definecolor{q}{RGB}{0,0,255}
\definecolor{r}{RGB}{80,255,80}
\newcommand{\highlight}[2]{
    \path [line cap=round, line join=round, line width=1.3em, draw = #1!20] #2;
}

\pgfdeclarelayer{background}
\pgfsetlayers{background,main}

\makeatletter
\pgfplotsset{
  cycle list/Dark2,
  cycle multiindex* list={
    mark list*\nextlist
    Dark2\nextlist
  },
  /pgfplots/flexlabels/.code n args={3}{%
    \pgfplotstableread[#3]{#1}\coordinate@table
    \pgfplotstablegetcolumn{#2}\of{\coordinate@table}\to\pgfplots@xticklabels
    \let\pgfplots@xticklabel=\pgfplots@user@ticklabel@list@x
  }
}
\makeatother

\newacro{NLP}{natural language processing}
\newacro{AA}{authorship attribution}
\newacro{TD}{topic detection}
\newacro{CTAA}{cross-topic authorship attribution}
\newacro{CGAA}{cross-genre authorship attribution}
\newacro{CLAA}{cross-language authorship attribution}
\newacro{CTTD}{cross-topic topic detection}
\newacro{CGTD}{cross-genre topic detection}
\newacro{CLTD}{cross-language topic detection}
\newacro{CLIR}{cross-language information retrieval}
\newacro{CLPA}{cross-language plagiarism analysis}

\title{DT-grams: Structured Dependency Grammar Stylometry for Cross-Language Authorship Attribution}
\toappear{}

\begin{document}
\numberofauthors{2}
\author{
    \alignauthor Benjamin Murauer\\
    \affaddr{Universität Innsbruck, Austria}\\
    \email{b.murauer@posteo.de}
    \alignauthor Günther Specht\\
    \affaddr{Universität Innsbruck, Austria}\\
    \email{guenther.specht@uibk.ac.at}
}
\maketitle
\begin{abstract}
Cross-language authorship attribution problems rely on either translation to enable the use of single-language features, or language-independent feature extraction methods.
Until recently, the lack of datasets for this problem hindered the development of the latter, and single-language solutions were performed on machine-translated corpora.
In this paper, we present a novel language-independent feature for authorship analysis based on dependency graphs and universal part of speech tags, called DT-grams (dependency tree grams), which are constructed by selecting specific sub-parts of the dependency graph of sentences.
We evaluate DT-grams by performing cross-language authorship attribution on untranslated datasets of bilingual authors, showing that, on average, they achieve a macro-averaged F1 score of 0.081 higher than previous methods across five different language pairs.
Additionally, by providing results for a diverse set of features for comparison, we provide a baseline on the previously undocumented task of untranslated cross-language authorship attribution.
\end{abstract}

\section{Introduction}

In cross-language authorship attribution, the true author of a previously unseen document must be determined from a set of candidate authors after training a model with documents from those candidates in a different language. 
Previous work in single-language attribution often relies on language-specific features.
Here, popular and powerful features often exploit character- and word-based measures \cite{Stamatatos2013,Eder2011}.
Using translation enables easy re-use of these features, and has been shown to be a useful tool for cross-language attribution \cite{bogdanova2014cross}.
However, setting up a custom machine translation system is an expensive operation in terms of time and resources.
From a scientific perspective, translations from commercial and therefore, closed-source systems are difficult to explain and reproduce, as the details of the models are unknown to the customer and commercial providers will likely try to improve their models, causing different translations of the same input over time.
Therefore, language-independent alternatives to traditional attribution features are crucial for cross-language attribution without translation.

Candidates for such features include high-level measurements like vocabulary or punctuation statistics \cite{Narayanan2012} or features that can be mapped to a general space like universal grammar representations \cite{bogdanova2014cross}.
In this paper, our first contribution is a novel type of classification feature, DT-grams (dependency tree grams), that is based on dependency graphs and universal part-of-speech (POS) tags, making it language-independent.
It calculates frequencies of substructures within a dependency graph similar to how in traditional n-grams, frequencies of character or word combinations in the original text are counted.
We show that this feature is efficient for cross-language authorship attribution, a problem in which documents of bilingual authors are classified, but the language differs between training and testing documents.
In our experiments, DT-grams outperform other approaches in this field consistently by an average F1\textsubscript{macro} score of 0.081. 

For the authorship attribution experiment, we use a dataset consisting of social media comments of bilingual authors in multiple language pairs.
This distinguishes this work from previous research, which used artificially constructed corpora due to the lack of data from multilingual authors \cite{bogdanova2014cross,llorens2016deep}. 
Thereby, classic novels by professional authors were used as training data, and human-translated versions of other novels by the same author are used as evaluation data. 
Although research has shown that human translation does not eliminate stylometric features \cite{venuti2017translator}, the original author still has only written in one language.
Therefore, we argue that the classification problem is, more strictly speaking, a translation obfuscation measurement rather than an authorship attribution problem.
By performing our evaluation experiments on the untranslated data from bilingual authors, we add a second contribution to this paper by providing the first baseline for true, untranslated cross-language authorship attribution.

Summarized, our contribution in this paper is twofold: (1) we present a new feature type DT-grams for cross-language authorship analysis, and (2) our evaluations represent a baseline for the novel problem of true, untranslated cross-language authorship attribution.
To ensure the reproducibility of our results, all of our data and code is published online\footnote{\url{https://git.uibk.ac.at/csak8736/gvdb2021-code}}.

\section{Related Work}
Cross-language authorship analysis is a significantly more difficult problem than its single-language version \cite{Stamatatos2013}, and in many cases, know-how learned from single-language authorship analysis can't be directly used.
For example, simple syntactic features like word or character n-grams are an effective feature for stylometry \cite{kestemont:2018}, but are not suitable when the training and testing documents only share a few words, or even characters when given a different alphabet.
Generally, using grammar features for authorship classification has been proven effective in many tasks ranging from attribution \cite{Luyckx2005,Zhang2018,Jafari2019} to plagiarism detection \cite{Tschuggnall2013a}.
Although these examples use language-specific grammar features in single-language settings, they show the general ability of these features to distinguish authorship, and language-independent grammar features such as universal POS tags allow for cross-language classification \cite{bogdanova2014cross}. 

Using different combinations of words by leveraging the dependency of sentences rather than the original word order has lead to increased classification performance \cite{Sidorov2013}.
However, this study does not make use of language-independent features but rather changes how word n-grams are constructed by providing an alternative measure of which words neighbor each other. 
Nevertheless, their findings suggest that the dependency relationships between words within sentences hold valuable information for authorship analysis.

Our proposed feature, DT-grams, leverages key findings of previous observations by combining language-independent universal POS tags in combination with dependency graphs.

Previous attempts at cross-language attribution define the task itself inconsistently and different approaches to this term are taken, including datasets of monolingual authors of different languages \cite{Stuart2013} or comparing the performance of feature families in mono-lingual attribution problems for different languages \cite{Eder2011}.
When refining the definition of cross-language attribution as the task of attributing authors that have written documents in multiple languages, and training and testing documents must be written in different languages, few existing studies remain:
\cite{bogdanova2014cross} use a variety of different features including the frequency of universal POS tags on attribution, but conclude that machine-translation followed by traditional attribution techniques provides the best results.
\cite{llorens2016deep} use differently sized windows in which vocabulary richness measurements are aggregated.
However, in both works, the datasets that were used contain human-translated novels, where the original author only wrote in one language and the source of the other languages was added by using translations of these works.
Although it has been shown that translation keeps stylistic features mostly intact \cite{venuti2017translator}, we claim that the setup by these studies more likely measures the extent to which the authorship was obfuscated by the translator rather than the authorship itself.
We state that authors writing in multiple languages are likely to do so in different styles, and we distinguish this problem as a different type of task.

Therefore, in this paper, we use social media texts that have been written by bilingual authors \cite{Murauer2019}.
While this change in text type makes it more difficult to compare the results directly to previous work, it also allows us to analyze a more comprehensive set of language pairs that are available within this resource, and have not been included in previous studies due to the lack of data.
More importantly though, by using this resource, our evaluations of the DT-grams feature along with several previously established baseline features provide first reference results for untranslated authorship attribution in five different language pairs.


\section{DT-grams Construction}
To construct the proposed DT-grams feature, we parse textual data to obtain dependency relationships between the words within sentences, which are then mapped to a tree structure.
Then, differently sized substructures are selected from those trees to produce sequences of DT-grams.
Finally, while some classification models used in our experiments use these sequences directly, we also reduce them to tf/idf-normalized frequencies to form a bag-of-DT-grams for other models used in the evaluation.
In the following section, these steps are explained in detail.

\subsection{Grammar Representations}
\label{sec:grammar}
\begin{figure}
    \Large
  \begin{tikzpicture}[every path/.style={-{Latex},out=90,in=90,looseness=1.5,shorten >=4mm}]
    \node (n1) {\begin{tabular}{c}the\\DET\end{tabular}};
    \node (n2) [right=1mm of n1] {\begin{tabular}{c}cat\\NOUN\end{tabular}};
    \node (n3) [right=1mm of n2] {\begin{tabular}{c}saw\\VERB\end{tabular}};
    \node (n4) [right=1mm of n3] {\begin{tabular}{c}a\\DET\end{tabular}};
    \node (n5) [right=1mm of n4] {\begin{tabular}{c}mouse\\NOUN\end{tabular}};
    \node (n6) [right=1mm of n5] {\begin{tabular}{c}in\\ADP\end{tabular}};
    \node (n7) [right=1mm of n6] {\begin{tabular}{c}the\\DET\end{tabular}};
    \node (n8) [right=1mm of n7] {\begin{tabular}{c}field\\NOUN\end{tabular}};

    \draw (n2.north) to node [above] {det} (n1.north) ;
    \draw (n3.north) to node [above] {nsubj} (n2.north) ;
    \draw (n3.north) to node [above] {dobj} (n5.north) ;
    \draw (n5.north) to node [above] {det} (n4.north) ;
    \draw (n5.north) to node [above] {nmod} (n8.north) ;
    \draw (n8.north) to node [above] {case} (n6.north) ;
    \draw (n8.north) to node [above] {det} (n7.north) ;

  \end{tikzpicture}

  \caption{Dependency graph representation of the sentence `the cat saw a mouse in the field'.}
  \label{fig:dependency_tree}
\end{figure}
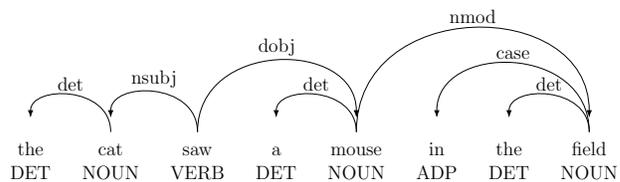

In the first step, the raw text is parsed by a dependency parser.
For this, we use the \textit{stanza}\footnote{\url{https://github.com/stanfordnlp/stanza}} python library.
This produces graphs as depicted in Figure~\ref{fig:dependency_tree}.
Along with the dependency graph, the parser also provides additional information for each word, including its lemma and universal POS tag.
The latter is a mapping from the more fine-grained language-dependent POS tag to a coarse, but language-independent universal tag \cite{nivre2016}, and we use it as a supplemental representation of the word itself and by discarding the original word.
This way, we construct a language-independent tree from the graph of each sentence, and encode both the relationship between the words as well as their grammatical role.

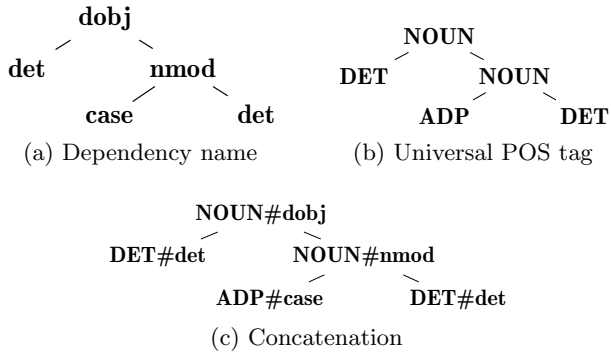
\begin{figure}
  \centering
  \subfloat[Dependency name\label{fig:tree_node_dep}]{
      \begin{forest}
    for tree = {l=0mm, l sep=1mm, s sep=1cm}
      [ \textbf{dobj}
        [\textbf{det}]
          [ \textbf{nmod}
            [ \textbf{case}]
            [ \textbf{det} ]
          ]
      ]
  \end{forest}
  }\hfill
  \subfloat[Universal POS tag\label{fig:tree_node_upos}]{
      \begin{forest}
    for tree = {l=0mm, l sep=1mm, s sep=1cm}
      [ \textbf{NOUN}
        [\textbf{DET}]
          [ \textbf{NOUN} 
            [ \textbf{ADP}]
            [ \textbf{DET}]
          ]
      ]
  \end{forest}
  }\\
  \subfloat[Concatenation\label{fig:tree_node_both}]{
      \begin{forest}
    for tree = {l=0mm, l sep=1mm, s sep=1cm}
      [ \textbf{NOUN\#dobj} 
        [\textbf{DET\#det}]
          [ \textbf{NOUN\#nmod} 
            [ \textbf{ADP\#case}]
            [ \textbf{DET\#det}]
          ]
      ]
  \end{forest}
  }
  \caption{Three node representations of the dependency graph of the subphrase "mouse in the field" from Figure~\ref{fig:dependency_tree} containing the name of the dependency (a), the universal POS tag (b), and both (c).}
  \label{fig:tree_node}
\end{figure}

We test three different representations of the nodes within the tree which are depicted in Figure~\ref{fig:tree_node}:
(1) the name of the incoming dependency (Figure~\ref{fig:tree_node_dep}), 
(2) the universal POS tag of the word (Figure~\ref{fig:tree_node_upos}), and
(3) both (Figure~\ref{fig:tree_node_both}).
This way, we hope to gain insight into which parts of the dependency graph are more important for authorship stylometry.
The resulting influence of these choices is discussed in Section~\ref{sec:results}.

A similar representation of sentences can be achieved by using constituency parsers, which we refrained from using for two reasons:
firstly, the availability of parser models for non-English languages is limited, and secondly, the resulting constituents are not language-independent and a global mapping must be used in order to perform cross-language classification.
While such mappings exist for POS tags \cite{nivre2016}, no similar resources for constituents are available to our knowledge.


\subsection{Tree Substructure Representations}
\label{sec:substructures}

\begin{figure}
  \centering
  \hspace{1mm}
  \subfloat[DT\textsubscript{anc} (ancestors)\label{fig:stripes}]{
    \begin{tikzpicture}
  \node at (-1,0) {\small blue=3};
  \node[tg](n0){} [sibling distance=3.5em]
    child {[sibling distance=1.5em] node[tg](n1){} 
      child{node[tg](n2){}}
      child{node[tg](n3){}}
    }
    child {[sibling distance=1.5em] node[tg](n4){} 
      child{node[tg](n5){}}
      child{node[tg](n6){}}
      child{node[tg](n7){}}
    }
    child {node[tg](n8){} 
      child{node[tg](n9){}}
    }
    ;

  \begin{pgfonlayer}{background}
    \highlight{blue}{(n0.center) -- (n4.center) -- (n5.center)}
  \end{pgfonlayer}
\end{tikzpicture}
  }%
  \hspace{1mm}
  \subfloat[DT\textsubscript{sib} (siblings)\label{fig:hline}]{
    \begin{tikzpicture}
  \node at (-1,0) {\small red=2};
  \node[tg](n0){} [sibling distance=3.5em]
    child {[sibling distance=1.5em] node[tg](n1){} 
      child{node[tg](n2){}}
      child{node[tg](n3){}}
    }
    child {[sibling distance=1.5em] node[tg](n4){} 
      child{node[tg](n5){}}
      child{node[tg](n6){}}
      child{node[tg](n7){}}
    }
    child {node[tg](n8){} 
      child{node[tg](n9){}}
    }
    ;

  \begin{pgfonlayer}{background}
    \highlight{red}{(n1.center) -- (n4.center)}
    \highlight{white}{(n9.center) -- (n7.center)}
    \highlight{white}{(n0.center) -- (n0.center)}
  \end{pgfonlayer}
\end{tikzpicture}
  }%
  \hspace{1mm}
  \subfloat[DT\textsubscript{pq} (PQ-grams)\label{fig:pqgram}]{
    \begin{tikzpicture}[
  sibling distance=2em,
  level distance=2em,]
  \node at (-0.75,0) {\small \begin{minipage}{2cm}blue=2\\ red=3\end{minipage}};
  \node[tg](n0){} [sibling distance=3.5em]
    child {[sibling distance=1.5em] node[tg](n1){} 
      child{node[tg](n2){}}
      child{node[tg](n3){}}
    }
    child {[sibling distance=1.5em] node[tg](n4){} 
      child{node[tg](n5){}}
      child{node[tg](n6){}}
      child{node[tg](n7){}}
    }
    child {node[tg](n8){} 
      child{node[tg](n9){}}
    }
    ;

  \begin{pgfonlayer}{background}
    \highlight{blue}{(n0.center) -- (n4.center)}
    \highlight{red}{(n5.center) -- (n7.center)}
    \highlight{white}{(n2.center) -- (n1.center)}
  \end{pgfonlayer}
\end{tikzpicture}
  }%
  \hspace{1mm}
  \subfloat[DT\textsubscript{inv} (inverted PQ) \label{fig:raingram}]{
    \begin{tikzpicture}
  \node at (-0.75,0) {\small \begin{minipage}{2cm}blue=2\\ red=3\end{minipage}};
  \node[tg](n0){} [sibling distance=3.5em]
    child {[sibling distance=1.5em] node[tg](n1){} 
      child{node[tg](n2){}}
      child{node[tg](n3){}}
    }
    child {[sibling distance=1.5em] node[tg](n4){} 
      child{node[tg](n5){}}
      child{node[tg](n6){}}
      child{node[tg](n7){}}
    }
    child {node[tg](n8){} 
      child{node[tg](n9){}}
    }
    ;

  \begin{pgfonlayer}{background}
    \highlight{blue}{(n1.center) -- (n2.center)}
    \highlight{blue}{(n1.center) -- (n3.center)}
    \highlight{blue}{(n4.center) -- (n5.center)}
    \highlight{blue}{(n4.center) -- (n6.center)}
    \highlight{blue}{(n4.center) -- (n7.center)}
    \highlight{blue}{(n8.center) -- (n9.center)}

    \highlight{red}{(n1.center) -- (n4.center) -- (n8.center)}
  \end{pgfonlayer}
\end{tikzpicture}
  }%
  \caption{DT-grams. Substructures are based on simple tree building blocks (a, b), PQ-grams by \cite{Tschuggnall2013a} (c) and an inverted form thereof (d).}
  \label{fig:substructures}
\end{figure}
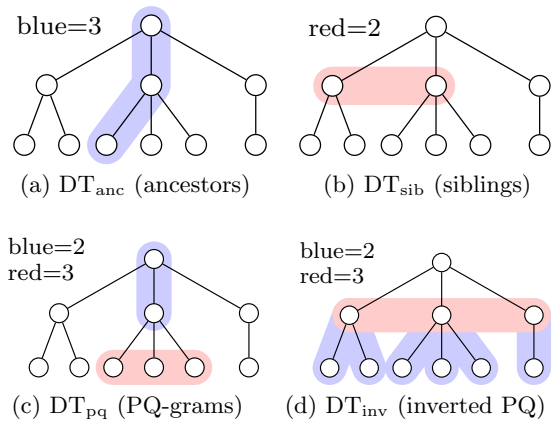

Along the lines of \cite{Tschuggnall2013a}, we use patterns of tree structures representing parts of the dependency tree. 
We propose several patterns, which we collectively call DT-grams and which are displayed in Figure~\ref{fig:substructures}.
The intention behind choosing these specific structures is as follows:
We first extract node combinations from direct ancestors (DT\textsubscript{anc}, Figure~\ref{fig:stripes}) and siblings (DT\textsubscript{sib}, Figure~\ref{fig:hline}), representing the most basic building blocks of a tree.
In Figure~\ref{fig:pqgram}, DT\textsubscript{pq} is displayed, based on the PQ-grams used by \cite{Tschuggnall2013a}.
Finally, we add DT\textsubscript{inv} that use a different order of sibling/ancestor relationship (Figure~\ref{fig:raingram}) compared to PQ-grams.

While character and word-based n-grams only have one dimension to scale (namely, n), these tree substructures can have more.
In general, two parameters control the number of siblings (red) and ancestors (blue) taken into account for each pattern, whereas DT\textsubscript{anc} and DT\textsubscript{sib} both only have one of those parameters each.
For DT\textsubscript{anc} and DT\textsubscript{sib}, setting the parameter to 1 results in calculating POS tag unigramsd.

To get instances of the DT-gram patterns from a tree, the substructure patterns are moved across the tree similar to a sliding-window, generating an instance of the substructure at every step.
Thereby, one has to define an order in which the DT-grams are parsed from the trees (i.e., depth-first or breadth-first).
If a substructure does not fit onto a certain position of a tree, the empty spots in the pattern are filled with a wildcard element \texttt{X}.
Thereby, an instance is generated for every step as long as at least one of the substructure's positions is filled with a non-wildcard node. 

This way, the sequence of DT-grams can either be used directly as input for a sequence-based model (e.g., a recurrent network), or the frequencies of the parsed instances can be used analogously to those of character or word n-grams.

For example, applying DT\textsubscript{anc} shown in Figure~\ref{fig:stripes} with its parameter set to 3 to the tree in Figure~\ref{fig:tree_node_dep} results in 11 substructures:
\textit{X-X-dobj},
\textit{X-dobj-det},
\textit{dobj-det-X},
\textit{det-X-X},
\textit{X-dobj-nmod},
\textit{dobj-nmod-case},
\textit{nmod-case-X},
\textit{case-X-X},
\textit{dobj-nmod-det},
\textit{nmod-det-X},
\textit{det-X-X},

Finally, the frequency of each produced instance is counted over the entire document, and these frequencies are then tf/idf-normalized over the entire dataset.

\section{Evaluation}

To evaluate the DT-grams feature, we perform cross-language authorship attribution using data from multiple language pairs and different classifiers, and we compare the results to different baseline features.


\begin{table}
  \setlength{\tabcolsep}{7pt}
  \centering 
\begin{tabular}{lcccc}
\toprule
  Languages &   A &  Docs & L\textsubscript{doc} & D/A\textsubscript{min} \\
\midrule
    EN + DE &  10 & 2,790 &                3,055 &                22 + 20 \\
 EN + DeepL &  10 & 2,790 &                3,055 &                22 + 20 \\
    EN + ES &  20 & 3,402 &                3,148 &                20 + 21 \\
    EN + PT &  37 & 4,481 &                2,996 &                20 + 20 \\
    EN + NL &  11 & 2,056 &                3,225 &                20 + 20 \\
    EN + FR &  45 & 7,374 &                3,142 &                21 + 20 \\
\bottomrule
\end{tabular}

\caption{Datasets used for evaluation.
    A denotes the number of authors.
    L\textsubscript{doc} denotes the average document length in characters.
    D/A\textsubscript{min} denotes the minimum number of documents written by each author in the respective languages in the first column.
    ``DeepL'' corresponds to the German documents machine-translated to English with DeepL.
  }
  \label{tab:datasets}
\end{table}

\subsection{Datasets}
Since there are no untranslated cross-language corpora available to our knowledge, we use the framework by \cite{Murauer2019} to generate several datasets by bilingual authors in different languages.
It collects user comments from the social media site Reddit and allows us to set minimum requirements for document count, length, and language.
We use this resource to evaluate the performance of DT-grams for different language pairs and generate bilingual datasets for the combinations presented in Table~\ref{tab:datasets}.
We choose five different language pairs which all contain English, which represents the largest portion of text in Reddit comments.
The other languages were chosen as they represent the largest non-English text sources for this corpus.
We set the parameters of the generation framework to produce corpora with at least 10 authors for each pair, where each author has at least 20 documents for both languages.
To increase the quality of the text documents, we also required a minimum document length of 3,000 characters.
The tools that generate these corpora perform preprocessing including replacing URLs with a tag \texttt{<URL>} or filtering messages that mainly consist of punctuation.
For a full list of preprocessing steps, we refer to the original publication by \cite{Murauer2019}.
We performed no additional preprocessing.
The resulting corpora are shown in Table~\ref{tab:datasets} and we provide them publicly for download\footnote{\url{https://git.uibk.ac.at/csak8736/gvdb2021-code}}.

In previous work, mono-lingual attribution techniques on machine-translated documents outperform cross-language techniques \cite{bogdanova2014cross}.
We therefore provide data to calculate such a baseline by using the commercial translation service DeepL\footnote{\url{https://www.deepl.com/}, translation performed in November 2019} to translate the German documents to English, creating a mono-lingual version of the German documents for comparison.
However, due to budgetary reasons, we only perform this step for one randomly picked language (German).

For each language pair $(A, B)$, we conduct all experiments both with training on $A$ and testing on $B$, as well as the other way around.

\subsection{Evaluation Strategy}
Since the parameterized datasets only define lower limits for the number of documents per author and the size of these documents, the resulting datasets have varying amounts of documents and authors.
We ensure that results from experiments using these datasets can be easily compared to each other by only selecting 10 random authors of each dataset, and selecting 10 random documents of each language from those authors.

To reduce bias, each of these evaluations is repeated 10 times, and the selected authors and documents are randomized in each repetition.
For each of these repetitions, all combinations of features and classifiers are tested, and the mean value of each combination across all repetitions is used as a representative for that combination.
This also functions as a supplement for traditional cross-validation, which is impossible for cross-domain classification as documents in the training set can't be used interchangeably for testing, which would break the cross-domain nature of the setup.
We are aware that this results in some datasets having a larger overlap between the repetitions than others, which is a flaw that might be mitigated in the future if more comprehensive corpora of bilingual authors become available, or direct comparison between results originating from differently sized datasets is not important.

\subsection{Models and Baselines}

\begin{figure}
  \centering
  \includegraphics[width=\linewidth]{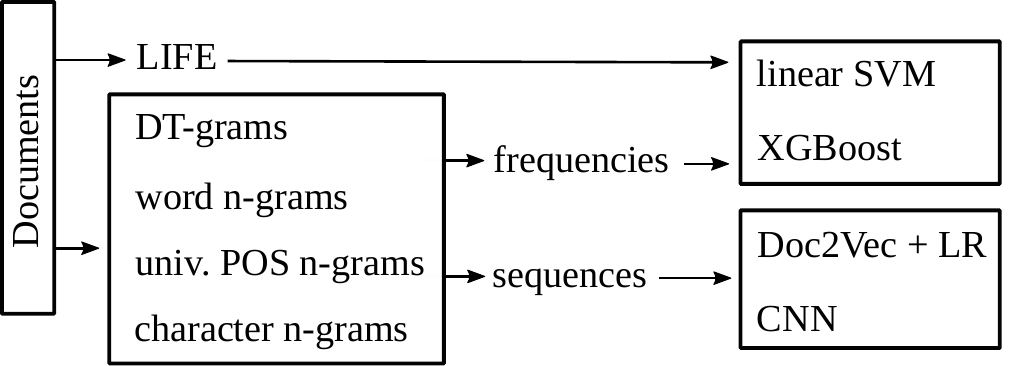}
  \caption{Models used in the experiments.}
  \label{fig:features_models}
\end{figure}

We test several different text classification models by following previous approaches in authorship attribution tasks.
These are summarized in Figure~\ref{fig:features_models}.

Firstly, calculating tf/idf-normalized frequencies of different types of n-grams has been used widely in the authorship analysis field, including character, word, or part-of-speech tag n-grams.
This approach can be used analogously by counting the frequencies of the parsed DT-grams and normalizing them using tf/idf.
We then test two commonly used classifiers: linear SVMs \cite{Stamatatos2013,Narayanan2012,Koppel2006} and extreme gradient boosting \cite{llorens2016deep}.
As comparison baselines of this category, we include results from character, word, and universal POS tag n-grams, whereby n ranges from 1 to 5.

Secondly, we utilize the Doc2Vec document embedding technique in combination with a logistic regression classifier, as proposed by \cite{gomez2018}.
For this solution, we have to define what a document is in terms of DT-grams, as their order is no longer well-defined.
We interpret each document as the sequence of DT-grams that is returned by the parser, which in our case uses a depth-first approach.
We include baselines for comparison along the lines of \cite{gomez2018}, which consist of character, word, and universal POS n-grams ranging from n=1 to 5.

Thirdly, we use a convolutional neural network proposed in \cite{Shrestha2017} by interpreting each DT-gram as a unique token used in the embedding layer of the network. 
Thereby, we use the same parameters and network layout as in \cite{Shrestha2017}, except for an increased embedding layer size to fit the larger documents.
We utilize the same depth-first order as in the second approach to define a sequence of tokens.
The baseline for this model uses character, word, and universal POS tag unigram representations of the documents.

As a further comparison baseline, we compute the vocabulary richness feature LIFE from \cite{llorens2016deep}, which counts the vocabulary frequency over differently sized windows and calculates various aggregated measures.
We refrain from using other language-agnostic features presented in related cross-language research \cite{bogdanova2014cross}, which depend on language-specific resources like sentiment databases, which are difficult to collect and even harder to compare.
Additionally, in their research, these approaches showed inferior performance compared to character-based features from machine-translated text.
We use the same linear SVM and extreme gradient boosting classifiers as the tf/idf frequency feature category to classify the documents with LIFE features (see Figure~\ref{fig:features_models}).

\begin{table}
  \centering
  \footnotesize
  \setlength{\tabcolsep}{4pt}
  \begin{tabular}{ll}
    \toprule
    Parameter & Values \\
    \midrule
    n-gram size & 1 - 3\\
    DT-gram structure & DT\textsubscript{anc}, DT\textsubscript{sib}, DT\textsubscript{pq}, DT\textsubscript{inv} \\
    DT-gram dim. sizes & 1 -- 4, 1 -- 4 \\
    C-value of SVM & 0.1, 1, 10 \\
    Doc2Vec emb. size & 50, 100, ..., 250 \\
    CNN batch size & 5, 10, 20 \\
    \bottomrule
  \end{tabular}
  \caption{Hyperparameters optimized by grid search. All n-gram sizes were tested individually for word, character and universal POS-tag n-grams.}
  \label{tab:parameters}
\end{table}

\section{Results and Discussion}
\label{sec:results}
We run the classification experiment for each model, each language pair in both directions, and every parameter combination shown in Table~\ref{tab:parameters}, generating an exhaustive grid of results.
In this section, different aggregations and selections of this entire result set are used to extract the key findings for this paper.

\subsection{Performance per Model}
\begin{table}

  \setlength{\tabcolsep}{3pt}
  \centering
  \subfloat[%
    Max. F1\textsubscript{macro} score of the models across all datasets.
    ``DeepL'' denotes the German documents machine-translated to English with DeepL.
    \label{tab:classifiers_languages}%
  ]{%
    \footnotesize
    \centering
\begin{tabular}{lcccccc}
\toprule
Model &             \sfrac{EN}{DE} &             \sfrac{EN}{ES} &             \sfrac{EN}{FR} &             \sfrac{EN}{NL} &             \sfrac{EN}{PT} &          \sfrac{EN}{DeepL} \\
\midrule
  svm & \textbf{ 0.375 }  & \textbf{ 0.291 }  & \textbf{ 0.310 }  & \textbf{ 0.277 }  & \textbf{ 0.246 }  & \textbf{ 0.479 }  \\
  xgb &             0.268 &             0.207 &             0.229 &             0.209 &             0.175 &             0.332 \\
  cnn &             0.112 &             0.108 &             0.104 &             0.102 &             0.119 &             0.133 \\
  d2v &             0.261 &             0.180 &             0.179 &             0.193 &             0.213 &             0.344 \\
\bottomrule
\end{tabular}
  }
  \setlength{\tabcolsep}{2pt}

  \subfloat[%
    Max. F1\textsubscript{macro} score of the  models across different features.
    \label{tab:classifiers_features}%
  ]{
    \footnotesize
    \centering
\begin{tabular}{lccccc}
\toprule
Model &              LIFE & \begin{minipage}{13mm}\centering Word\\ n-grams\end{minipage} & \begin{minipage}{13mm}\centering Char.\\ n-grams\end{minipage} & \begin{minipage}{14mm}\centering Uni. POS\\ n-grams\end{minipage} &          DT-grams \\
\midrule
  svm &             0.110 &                                  \textbf{ 0.396 }  &                                  \textbf{ 0.479 }  &                                  \textbf{ 0.385 }  & \textbf{ 0.453 }  \\
  xgb & \textbf{ 0.157 }  &                                              0.189 &                                              0.332 &                                              0.282 &             0.328 \\
  cnn &                 - &                                              0.092 &                                              0.075 &                                              0.133 &             0.102 \\
  d2v &                 - &                                              0.143 &                                              0.341 &                                              0.344 &             0.336 \\
\bottomrule
\end{tabular}

  }
  \caption{F1\textsubscript{macro} of the models across different datasets (a) and features (b).
  }
  \label{tab:classifiers}
\end{table}
Table~\ref{tab:classifiers} shows that the linear support vector machine with tf/idf frequency features outperforms all other models in every language combination and for most of the feature categories.
In the case of the vocabulary richness feature LIFE, we can confirm the results of the original work that the random forest-based approach outperforms the support vector machine \cite{llorensDissertation}.

We suspect that the CNN model underperforms because we have significantly less training documents than in the original paper, in which case network models have been shown to have trouble capturing the style of authors \cite{kestemont:2018}.

While the document embedding model (d2v in the table) outperforms the frequency-based features with the extreme boosting trees in some cases, it does not reach the support vector machine's F1 scores in any language or feature set.

\subsection{Performance per Feature Category}
\pgfplotstableread[col sep=comma] {generated_assets/features_overview.csv}\featurestable
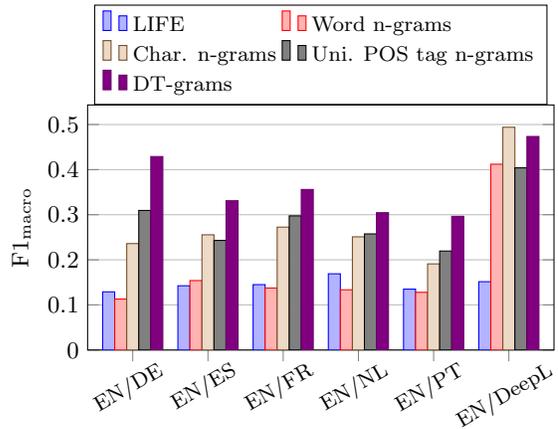
\begin{figure}
  \centering
  \begin{tikzpicture}
    \begin{axis}[
        ybar=0pt,  
        ylabel=F1\textsubscript{macro},
        ylabel near ticks,
        ymin=0, 
        x=10mm,  
        y=6mm/0.1,
        enlarge x limits={abs=6mm},  
        bar width= 1.6mm,
        flexlabels={generated_assets/features_overview.csv}{language}{col sep=comma},
        xtick=data,
        xtick pos=left,
        ymajorgrids=true,
        xticklabel style={text height=0.1ex, rotate=30, font=\small}, 
        ytick={0.0, 0.1,...,0.6},
        legend columns=2,
        legend cell align={left},
        legend style={
          font=\small,
          anchor=south,
          at={(.5,1)}
        }
      ]
      \addplot table [x expr=\coordindex, y=life, col sep=comma] {\featurestable}; 
      \addlegendentry{LIFE}
      \addplot table [x expr=\coordindex, y=word_ngrams, col sep=comma] {\featurestable}; 
      \addlegendentry{Word n-grams}
      \addplot table [x expr=\coordindex, y=char_ngrams, col sep=comma] {\featurestable}; 
      \addlegendentry{Char. n-grams }
      \addplot table [x expr=\coordindex, y=upos_ngrams, col sep=comma] {\featurestable};
      \addlegendentry{Uni. POS tag n-grams}
      \addplot table [x expr=\coordindex, y=dtgrams, col sep=comma] {\featurestable}; 
      \addlegendentry{DT-grams}
 
    \end{axis}
  \end{tikzpicture}
  \caption{Comparison of the highest F1\textsubscript{macro} scores for different feature types.
The different datasets are plotted on the x-axis, where ``DeepL'' stands for the documents that have been machine-translated from German to English.
For layout reasons, experiments that differ only in classification direction (e.g., en $\to$ de and de $\to$ en) are averaged, whereas the difference in F1\textsubscript{macro} between the directions was below 0.02 for each pair.
The DT-gram feature outperforms the next best feature by 0.081 F1\textsubscript{macro} averaged over all untranslated language pairs.
  }
  \label{fig:result_features}
\end{figure}

\begin{table}
  \centering
  \footnotesize
  \setlength{\tabcolsep}{3pt}
\begin{tabular}{lcccccc}
\toprule
DT\textsubscript{g} &            \sfrac{EN}{DE} &            \sfrac{EN}{ES} &            \sfrac{EN}{FR} &            \sfrac{EN}{NL} &            \sfrac{EN}{PT} &         \sfrac{EN}{DeepL} \\
\midrule
 DT\textsubscript{anc} &             0.33 &             0.21 &             0.23 &             0.23 &             0.18 &             0.42 \\
 DT\textsubscript{sib} &             0.29 &             0.24 &             0.24 &             0.25 &             0.25 &             0.42 \\
  DT\textsubscript{pq} &             0.35 &             0.26 &             0.28 & \textbf{ 0.28 }  & \textbf{ 0.29 }  & \textbf{ 0.43 }  \\
 DT\textsubscript{inv} & \textbf{ 0.37 }  & \textbf{ 0.30 }  & \textbf{ 0.29 }  &             0.23 &             0.27 &             0.43 \\
\bottomrule
\end{tabular}

  \caption{Max. F1\textsubscript{macro} score of each DT-gram type.} 
  \label{tab:treegram_shapes}
\end{table}

\pgfplotstableread[col sep=comma] {generated_assets/p_de.csv}\tabpde
\pgfplotstableread[col sep=comma] {generated_assets/p_deepL.csv}\tabpdeepL
\pgfplotstableread[col sep=comma] {generated_assets/p_es.csv}\tabpes
\pgfplotstableread[col sep=comma] {generated_assets/p_fr.csv}\tabpfr
\pgfplotstableread[col sep=comma] {generated_assets/p_nl.csv}\tabpnl
\pgfplotstableread[col sep=comma] {generated_assets/p_pt.csv}\tabppt

\pgfplotstableread[col sep=comma] {generated_assets/q_de.csv}\tabqde
\pgfplotstableread[col sep=comma] {generated_assets/q_deepL.csv}\tabqdeepL
\pgfplotstableread[col sep=comma] {generated_assets/q_es.csv}\tabqes
\pgfplotstableread[col sep=comma] {generated_assets/q_fr.csv}\tabqfr
\pgfplotstableread[col sep=comma] {generated_assets/q_nl.csv}\tabqnl
\pgfplotstableread[col sep=comma] {generated_assets/q_pt.csv}\tabqpt

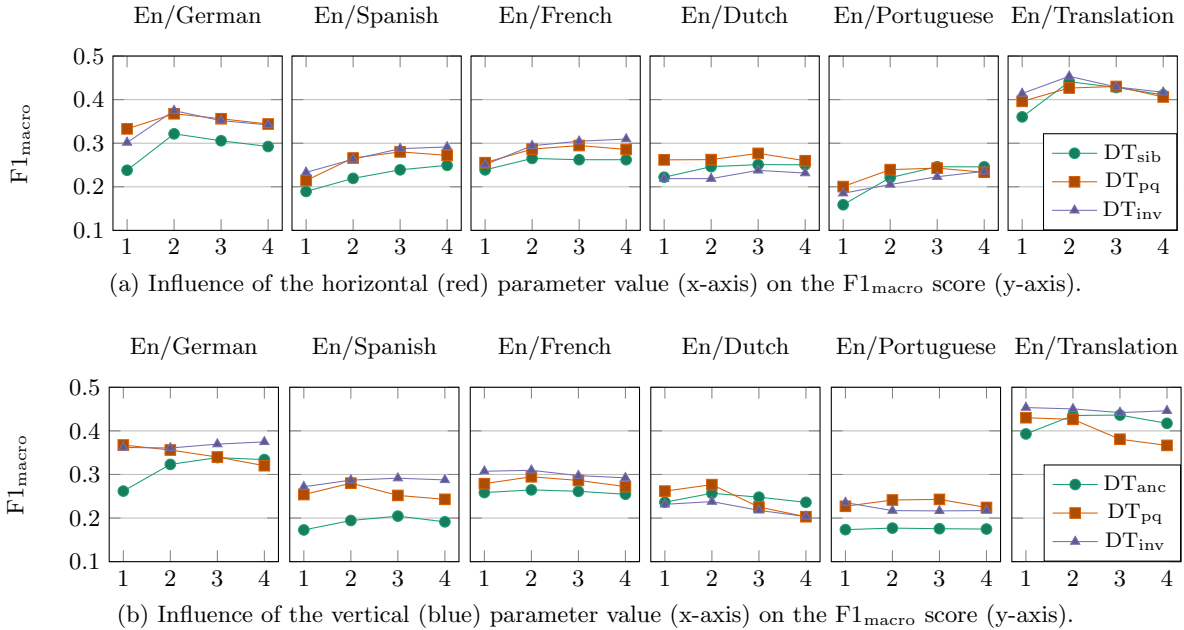
\begin{figure*}
  \centering
  \subfloat[Influence of the horizontal (red) parameter value (x-axis) on the F1\textsubscript{macro} score (y-axis).\label{fig:red}]{%
  \begin{tikzpicture}
    \begin{groupplot}[
        group style={
          group size=6 by 1,           
          horizontal sep=0.13cm,
          ylabels at=edge left,
          yticklabels at=edge left,
        },
        ylabel={F1\textsubscript{macro}},
        ymin=0.1,
        ymax=0.5,
        x=1cm/1.6,
        y=5.8mm/0.1,
        ymajorgrids=true,
        ytick={0.0, 0.1,...,0.55},
        legend style={
          font=\small,
          anchor=south east,
          at={(1,0)}
        }
      ]
      \nextgroupplot[title=En/German\strut]
      \addplot table [x=P, y=TS2] {\tabpde};
      \addplot table [x=Q, y=TS3] {\tabqde};
      \addplot table [x=P, y=TS4] {\tabpde};

      \nextgroupplot[title=En/Spanish\strut]
      \addplot table [x=P, y=TS2] {\tabpes};
      \addplot table [x=Q, y=TS3] {\tabqes};
      \addplot table [x=P, y=TS4] {\tabpes};

      \nextgroupplot[title=En/French\strut]
      \addplot table [x=P, y=TS2] {\tabpfr};
      \addplot table [x=Q, y=TS3] {\tabqfr};
      \addplot table [x=P, y=TS4] {\tabpfr};

      \nextgroupplot[title=En/Dutch\strut]
      \addplot table [x=P, y=TS2] {\tabpnl};
      \addplot table [x=Q, y=TS3] {\tabqnl};
      \addplot table [x=P, y=TS4] {\tabpnl};

      \nextgroupplot[title=En/Portuguese\strut]
      \addplot table [x=P, y=TS2] {\tabppt}; 
      \addplot table [x=Q, y=TS3] {\tabqpt}; 
      \addplot table [x=P, y=TS4] {\tabppt};

      \nextgroupplot[title=En/Translation\strut]
      \addplot table [x=P, y=TS2] {\tabpdeepL};\addlegendentry{DT\textsubscript{sib}};
      \addplot table [x=Q, y=TS3] {\tabqdeepL};\addlegendentry{DT\textsubscript{pq}};
      \addplot table [x=P, y=TS4] {\tabpdeepL};\addlegendentry{DT\textsubscript{inv}};
    \end{groupplot}
    \end{tikzpicture}
  }

  \subfloat[Influence of the vertical (blue) parameter value (x-axis) on the F1\textsubscript{macro} score (y-axis).\label{fig:blue}]{%
  \begin{tikzpicture}
    \begin{groupplot}[
        group style={
          group size=6 by 1,           
          horizontal sep=0.15cm,
          ylabels at=edge left,
          yticklabels at=edge left,
        },
        ylabel={F1\textsubscript{macro}},
        ymin=0.1,
        ymax=0.5,
        x=1cm/1.6,
        y=5.8mm/0.1,
        ymajorgrids=true,
        ytick={0.0, 0.1,...,0.55},
        legend style={
          font=\small,
          anchor=south east,
          at={(1,0)}
        }
      ]
      \nextgroupplot[title=En/German\strut]
      \addplot table [x=P, y=TS1] {\tabpde}; 
      \addplot table [x=P, y=TS3] {\tabpde};
      \addplot table [x=Q, y=TS4] {\tabqde};

      \nextgroupplot[title=En/Spanish\strut]
      \addplot table [x=P, y=TS1] {\tabpes};
      \addplot table [x=P, y=TS3] {\tabpes};
      \addplot table [x=Q, y=TS4] {\tabqes};

      \nextgroupplot[title=En/French\strut]
      \addplot table [x=P, y=TS1] {\tabpfr};
      \addplot table [x=P, y=TS3] {\tabpfr};
      \addplot table [x=Q, y=TS4] {\tabqfr};

      \nextgroupplot[title=En/Dutch\strut]
      \addplot table [x=P, y=TS1] {\tabpnl};
      \addplot table [x=P, y=TS3] {\tabpnl};
      \addplot table [x=Q, y=TS4] {\tabqnl};

      \nextgroupplot[title=En/Portuguese\strut]
      \addplot table [x=P, y=TS1] {\tabppt}; 
      \addplot table [x=P, y=TS3] {\tabppt}; 
      \addplot table [x=Q, y=TS4] {\tabqpt}; 

      \nextgroupplot[title=En/Translation\strut]
      \addplot table [x=P, y=TS1] {\tabpdeepL};\addlegendentry{DT\textsubscript{anc}};
      \addplot table [x=P, y=TS3] {\tabpdeepL};\addlegendentry{DT\textsubscript{pq}};
      \addplot table [x=Q, y=TS4] {\tabqdeepL};\addlegendentry{DT\textsubscript{inv}};
    \end{groupplot}
  \end{tikzpicture}
  }
  \caption{%
    Influence of the horizontal (a) and vertical (b) DT-gram parameter sizes.
    Note that DT\textsubscript{sib} is only included in (a) as it lacks a vertical parameter, and likewise, DT\textsubscript{anc} is only included in (b).}
  \label{fig:treegram_sizes}
\end{figure*}

Figure~\ref{fig:result_features} displays the highest F1\textsubscript{macro} score for each frequency feature category and dataset.
It becomes clear that the vocabulary richness feature LIFE is not able to model the authors effectively.
An explanation for this is found in the basic principle behind the feature itself, which counts aggregated vocabulary richness measures across sliding windows over the document.
Being originally developed for classifying entire novels from professional authors allowed these window sizes to be large and carry more information than is the case with shorter texts.
Likewise and unsurprisingly, the word n-grams are not able to model authorship except for the machine-translated dataset, which is the only case where a significant intersection between training and validation vocabulary can be expected.

Confirming the results of \cite{bogdanova2014cross}, we observe that traditional features are effective in classifying machine-translated text, outperforming all other features. 
We can also confirm their finding that machine-translation increases the performance of language-independent features.
Interestingly, the character n-gram features perform well above the 10\% random baseline also for the non-translated datasets.
This suggests a measure of similarity between these languages, but we leave the interpretation of these results to the field of linguistics.
Future experiments including datasets from less related language families such as Japanese or Arabic may provide further insights into this relationship.

The proposed DT-gram feature is the most effective feature for the untranslated scenarios, outperforming the next best feature across the language pairs by an average of 0.081 F1\textsubscript{macro}.

This suggests that the grammatical characteristics of multilingual authors are kept across languages.
The performance of these features consistently outperforms n-grams constructed from the universal POS tag-based on the original word order, we conclude that the dependency relationships between the words and therefore, a grammatical style contribute to an author's stylometric fingerprint.

When comparing the different languages, we can see a clear difference in classification performance. 
For the two grammatical feature types, namely universal POS tag n-grams and DT-grams, the results of the German dataset show better F1 scores compared to the other languages. 
One possible explanation for this result the overall higher grammar complexity of German compared to the other languages \cite{sadeniemi2008}, which would, in turn, suggest that either (1) classification across languages with grammars of different complexity, or (2) classification across languages with general high complexity improve the usefulness of grammar features themselves.

However, to answer these questions, additional language combinations must be analyzed, which may prove difficult for low-resource languages given the already small amount of available data from bilingual authors for languages that are not considered low-resource.



In summary, no approach is able to beat traditional methods performed on machine-translated texts, but our proposed DT-gram feature outperforms all other tested features on untranslated cross-language scenarios, especially on German documents.
It represents a promising start for future development and research of true cross-language authorship attribution.

\subsection{Performance by Tree Node Structure}
\begin{table}
  \centering
  \footnotesize
  \setlength{\tabcolsep}{3pt}
\begin{tabular}{ccccccc}
\toprule
Node &             \sfrac{EN}{DE} &             \sfrac{EN}{ES} &             \sfrac{EN}{FR} &             \sfrac{EN}{NL} &             \sfrac{EN}{PT} &          \sfrac{EN}{DeepL} \\
\midrule
  Dep. &             0.366 &             0.239 &             0.274 &             0.257 &             0.218 &             0.445 \\
 U.POS & \textbf{ 0.375 }  & \textbf{ 0.291 }  & \textbf{ 0.310 }  & \textbf{ 0.277 }  & \textbf{ 0.246 }  &             0.450 \\
  both &             0.368 &             0.232 &             0.294 &             0.262 &             0.235 & \textbf{ 0.453 }  \\
\bottomrule
\end{tabular}
  \caption{Max. F1\textsubscript{macro} scores of different internal node layouts for the dependency tree.}
  \label{tab:tree_node}
\end{table}
As described in Section~\ref{sec:grammar}, we tried different representations of the internal nodes of the dependency tree structure.
In Table~\ref{tab:tree_node}, the best results for each of these can be found.
Interestingly, the type of dependency which is used in the graph does not seem to have a large impact on the classification performance, but rather using only the structure of the graph along with the universal POS tag of each word shows the biggest advantage.

\subsection{Tree Substructure Performance Analysis}

As we demonstrated the general efficiency of the dependency tree-based features, Table~\ref{tab:treegram_shapes} shows how the different DT-grams perform on each language combination.
In general, the substructures that combine ancestor and sibling nodes (DT\textsubscript{pq} and DT\textsubscript{inv}) outperform the more simple patterns for each language and suggest that complex structures in grammatical style are a valuable stylometric feature for bilingual authors across languages.

Figure~\ref{fig:treegram_sizes} shows a more detailed analysis of how the sizes of the two parameters influence this result. 
For both the vertical and horizontal parameters, the optimal value is between 2 and 3, depending on the language and substructure, which is similar to reported optimal values for character n-grams \cite{Stamatatos2013}.
Only DT\textsubscript{anc} benefits from a higher vertical parameter size, especially in German documents, which may benefit from even higher values of the respective parameter.
While Spanish shows the least difference in classification performance across the different parameter sizes, it is difficult to draw conclusions from the other languages, indicating that more data is required for further experiments.

\section{Conclusion}

In this paper, we have presented a novel type of classification feature called DT-grams, based on dependency graphs and universal POS tags.
We have shown in experiments that DT-grams able to efficiently model stylometric fingerprints of bilingual authors across languages, premiering authorship analysis even in cases where machine-translation is unavailable, with an average lead of 0.081 F1\textsubscript{macro} to the next best approach tested in our experiments.
Additionally, we have expanded the field of cross-language authorship attribution by providing baseline results for the previously undocumented problem of untranslated cross-language authorship attribution of bilingual authors and analyzed results of 5 different language pairs.
Finally, we have collected findings including unexpectedly good performances of language-dependent features applied to cross-language settings as well as significant differences across language pairs.

The most important limitations of our approach are the dependency on the performance of the external parsing tools used, which may differ in quality across languages, as well as the superior performance of approaches based on machine-translation.

In future work, we want to investigate on using more specialized syntax classification models like tree-LSTMs \cite{Tai2015} or more complex syntactic networks \cite{Jafari2019}, as well as combining multiple feature categories to further improve classification results in both cross- and single-language experiment settings.


\bibliographystyle{abbrv}
\bibliography{lit}

\end{document}